\documentclass[review]{elsarticle}

\usepackage{lineno,hyperref}
\modulolinenumbers[5]

\journal{Journal of \LaTeX\ Templates}

\usepackage{tabularx}
\usepackage{multirow}
\begin{document}

\begin{frontmatter}

\title{A Comprehensive Survey of Image Augmentation Techniques for Deep Learning}
% \tnotetext[mytitlenote]{Fully documented templates are available in the elsarticle package on \href{http://www.ctan.org/tex-archive/macros/latex/contrib/elsarticle}{CTAN}.}

%% Group authors per affiliation:
\author{Mingle Xu}
\address{Department of Electronics Engineering, Jeonbuk National University, Jeonbuk 54896, South Korea}

\author{Sook Yoon\corref{mycorrespondingauthor}}
%\fnref{myfootnote}
\cortext[mycorrespondingauthor]{Corresponding author}
\address{Department of Computer Engineering, Mokpo National University, Jeonnam 58554, South Korea}

\author{Alvaro Fuentes}
\address{Core Research Institute of Intelligent Robots, Jeonbuk National University, Jeonbuk 54896, South Korea}

\author{Dong Sun Park\corref{mycorrespondingauthor}}
%\fnref{myfootnote}
\address{Core Research Institute of Intelligent Robots, Jeonbuk National University, Jeonbuk 54896, South Korea}

\begin{abstract}
Although deep learning has achieved satisfactory performance in computer vision, a large volume of images is required. However, collecting images is often expensive and challenging. Many image augmentation algorithms have been proposed to alleviate this issue. Understanding existing algorithms is, therefore, essential for finding suitable and developing novel methods for a given task. In this study, we perform a comprehensive survey of image augmentation for deep learning using a novel informative taxonomy. To examine the basic objective of image augmentation, we introduce challenges in computer vision tasks and vicinity distribution. The algorithms are then classified among three categories: model-free, model-based, and optimizing policy-based. The model-free category employs the methods from image processing, whereas the model-based approach leverages image generation models to synthesize images. In contrast, the optimizing policy-based approach aims to find an optimal combination of operations. Based on this analysis, we believe that our survey enhances the understanding necessary for choosing suitable methods and designing novel algorithms.
\end{abstract}


\end{frontmatter}

% \linenumbers

\section{Introduction}
Over the recent years, deep learning has achieved significant improvements in computer vision based on three key elements, efficient computing devices, powerful algorithms, and large volumes of image. A main work over the last decade was designing a powerful model with numerous trainable parameters\footnote{https://spectrum.ieee.org/andrew-ng-data-centric-ai}. The training of such a model requires a large volume of image to achieve competitive performance. However, collecting image is frequently an expensive and challenging process. Obtaining satisfactory performance with a limited dataset is particularly challenging in practical applications, such as medical \cite{litjens2017survey} and agricultural images \cite{100}.

To address this issue, image augmentation has been confirmed to be an effective and efficient strategy \cite{1, 2}.
As listed in Table \ref{table1}, many image augmentation methods have been utilized for image classification and object detection. Understanding existing image augmentation methods is, therefore, crucial in deploying suitable algorithms. Although similar surveys have been conducted previously \cite{8, 9, 10}, our study is characterized by several essential differences. First, we do not confine ourselves to a specific type of image, such as facial images \cite{4}. Likewise, we consider many types of image augmentation algorithms, including generative adversarial networks \cite{6} and image mixing \cite{7}. Third, we do not focus on a specific application, such as object detection \cite{8}. Conversely, we consider image classification and object detection as two primary applications, along with other image and video applications such as segmentation and tracking. Finally, unlike two related studies \cite{9, 10}, our survey encompasses more recent yet effective image augmentation algorithms such as instance level multiple image mixing, as well as comprehensive analysis of model-based methods. Consequently, this paper encompasses a wider range of algorithms that yield a novel informative taxonomy.

Specifically, we first explain why different image augmentation algorithms have been designed and leveraged across diverse applications. More specifically, challenges in computer vision and vicinity distribution are introduced to illustrate the necessity of image augmentation. By augmenting image data, the aforementioned challenges can be mitigated, and the vicinity distribution space can be dilated, thereby improving trained model's generalizability. Based on this analysis, we argue that novel image augmentation methods are promising when new challenges are recognized. Simultaneously, once a challenge is observed in an application, it can be mitigated using an appropriate augmentation method.

In summary, our study makes the following contributions.
\begin{itemize}
    \item We examine \emph{challenges} and \emph{vicinity distribution} to demonstrate the necessity of image augmentation for deep learning.
    \item We present a comprehensive survey on image augmentation with a novel \emph{informative taxonomy} that encompasses a wider range of algorithms.
    \item We discuss the current situation and future direction for image augmentation, along with three relevant topics: understanding image augmentation, new strategy to leverage image augmentation, and feature augmentation.
\end{itemize}

The reminder of this paper is organized as follows. The second section introduces the research taxonomy. We then present two basic inspiration of image augmentation in the third section: the challenges of computer vision tasks and the vicinity distribution. Model-free image augmentation is covered in the fourth section, whereas the model-based methods are discussed in the fifth section. The process of determining an optimal image augmentation is introduced in the six section, followed by a discussion section. Concluding remarks are presented in the final section.

\begin{table}[h!]
\begin{center}
\scriptsize
\begin{tabularx}{\textwidth}{|r|X|}
\hline
Paper & Image augmentation method \\
\hline
AlexNet \cite{27} & Translate, Flip, Intensity Changing \\
% VGGNet \cite{30} & Translate, Flip, Crop, Intensity changing \\
% GoogLeNet \cite{28} & Crop, Elastic distortion \\
ResNet \cite{34} & Crop, Flip \\
% DRN \cite{120} & Crop, Flip, Intensity Changing \\
% PyramidNet \cite{121} & Flip, Translate \\
DenseNet \cite{35} & Flip, Crop, Translate \\
% ResNext \cite{122} & Translate, Flip \\
MobileNet \cite{123} & Crop, Elastic distortion \\
% DPN \cite{124} & Random noise, Crop, Flip \\
% ShuffleNet \cite{125} & Translate, Flip \\
NasNet \cite{126} & Cutout, Crop, Flip \\
% SENet \cite{36} & Crop, Flip \\
% EfficientNet \cite{127} & AutoAugment \\
ResNeSt \cite{145} & AutoAugment, Mixup, Crop \\
% GhostNet \cite{146} & Crop, Flip \\
% EfficientNetv2 \cite{128} & RandAugment, Mixup, Cutout \\
DeiT \cite{129} & AutoAugmentat, RandAugment, Random Erasing, Mixup, CutMix \\
Swin Transformer \cite{130} & RandAugment, Mixup, CutMix, Random Erasing \\
% MLP-Mixer \cite{131} & RandAugment, Mixup \\
% ResMLP \cite{132} & AutoAugment, RandAugment, Random Erasing, Mixup, CutMix \\
% SPACH \cite{134} & RandAugment, Random Erasing, Mixup, CutMix \\
% ConvMixer \cite{133} & RandAugment, Mixup, CutMix, Random Erasing \\
\hline
% SPPNet \cite{136} & Crop, Flip \\
Faster R-CNN \cite{138} & Flip \\
YOLO \cite{139} & Scale, Translate, Color space \\
SSD \cite{140} & Crop, Resize, Flip, Color Space, Distortion \\
% YOLOv2 \cite{141} & Crop, Rotate, Color space \\
YOLOv4 \cite{50} & Mosaic, Distortion, Scale, Color space, Crop, Flip, Rotate, Random erase, Cutout, Hide-and-Seek, GridMask, Mixup, CutMix, StyleGAN \\
% \multirow[t]{3}{*}{YOLOv4 \cite{50}} & Mosaic, Distortion, Scale, Color space, \\
%     & Crop, Flip, Rotate, Random erase, Cutout, Hide-and-Seek, \\
%     & GridMask, Mixup, CutMix, StyleGAN \\
\hline
\end{tabularx}
\end{center}
\caption{Image augmentation algorithms used studies pertaining to image classification (up) and object detection (bottom).}
\label{table1}
\end{table}

\section{Taxonomy}
% \begin{figure}[t]
% \begin{center}
% \includegraphics[width=1.0\linewidth]{Image+Augmentation.eps}
% \end{center}
% \caption{Taxonomy of image augmentation.}
% \label{fig6}
% \end{figure}

As shown in Table \ref{tab:taxonomy}, we classify the image augmentation algorithms among three main categories. A model-free approach does not utilize a pre-trained model to perform image augmentation, and may use single or multiple images. Conversely, model-based algorithms require the image augmentation algorithms to generate images using trained models. The augmentation process may unconditional, label-conditional, or image-conditional. Finally, Optimizing policy-based algorithms determine the optimal operations with suitable parameters from a large parameter space. These algorithms can further be sub-categorized into reinforcement learning-based and adversarial learning-based method. The former leverages a massive search space consisting of diverse operations and their magnitudes, along with an agent to find the optimal policy within the search space. In contrast, adversarial learning-based methods locate algorithms with the corresponding magnitude to allow the task model to have a sufficiently large loss.

\begin{table}[h!]
    \centering
    \scriptsize
    \begin{tabularx}{\textwidth}{|p{0.15\textwidth}|p{0.20\textwidth}|p{0.15\textwidth}|X|}
        \hline
        \multicolumn{3}{|c|}{Categories} & Relevant methods \\
        \hline
        \multirow{5}{*}{Model-free} & \multirow{3}{*}{Single-image} & Geometrical transformation & translation, rotation, flip, scale, elastic distortion. \\
        \cline{3-4}
        & & Color image processing & jittering.  \\
        \cline{3-4}
        & & Intensity transformation & blurring and adding noise, Hide-and-Seek \cite{40}, Cutout \cite{41}, Random Erasing \cite{42}, GridMask \cite{43}. \\
        \cline{2-4}
        & \multirow{2}{*}{Multiple-image} & Non-instance-level & SamplePairing \cite{47}, Mixup \cite{23}, BC Learning \cite{48}, CutMix \cite{49}, Mosaic \cite{50}, AugMix \cite{53}, PuzzleMix \cite{51}, Co-Mixup \cite{kim2020co}, SuperMix \cite{52}, GridMix \cite{54}. \\
        \cline{3-4}
        & & Instance-level & CutPas \cite{56}, Scale and Blend \cite{57}, Context DA \cite{58}, Simple CutPas \cite{59}, Continuous CutPas \cite{60}. \\
        
        \hline
        \multirow{4}{*}{Model-based} & \multicolumn{2}{l|}{Unconditional} & DCGAN \cite{75}, \cite{76, 77, 78} \\
        \cline{2-4}
        & \multicolumn{2}{l|}{Label-conditional} & BDA \cite{79}, ImbCGAN \cite{84}, BAGAN \cite{83}, DAGAN \cite{86}, MFC-GAN \cite{82}, IDA-GAN \cite{85}. \\
        \cline{2-4}
        & \multirow{2}{*}{Image-conditional} & Label-preserving & S+U Learning \cite{92}, AugGAN \cite{91}, Plant-CGAN \cite{93}, StyleAug \cite{97}, Shape bias \cite{geirhos2018imagenet}. \\
        \cline{3-4}
        & & Label-changing & EmoGAN \cite{99}, $\delta$-encoder \cite{102}, Debiased NN \cite{103}, StyleMix \cite{104}, GAN-MBD \cite{101}, SCIT \cite{100}. \\
        
        \hline
        Optimizing policy-based & \multicolumn{2}{l|}{Reinforcement learning-based} & AutoAugment \cite{105}, Fast AA \cite{107}, PBA \cite{109}, Faster AA \cite{108}, RandAugment \cite{106}, MADAO \cite{110}, LDA \cite{114}, LSSP \cite{113}. \\
        \cline{2-4}
         & \multicolumn{2}{l|}{Adversarial learning-based} & ADA \cite{116}, CDST-DA \cite{117}, AdaTransform \cite{115}, Adversarial AA \cite{111}, IF-DA \cite{119}, SPA \cite{118}. \\
        \hline
    \end{tabularx}
    \caption{Taxonomy with relevant methods.}
    \label{tab:taxonomy}
\end{table}

\section{Motivation to perform image augmentation}
% In this section, we begin with the challenges and vicinity distribution to understand the necessity to perform image augmentation for deep learning.

\subsection{Challenges}

\begin{table}[h!]
\scriptsize
\begin{tabularx}{\textwidth}{ |p{0.2\textwidth}|p{0.3\textwidth}|X| }
  \hline
  Challenges & Descriptions & Strategies and related studies \\
  \hline
  \noalign{\vskip 0.2mm}
  \hline
  
  Images variations & The following basic variations exist in many datasets and applications, including illumination, deformation, occlusion, background, viewpoint, and multiscale, as shown in Figure \ref{fig1}. 
  & Geometrical transformation and color image processing improve the majority of the variations. Occlusion: Hide-and-Seek \cite{40}, Cutout \cite{41}, Random Erasing \cite{42}, GridMask \cite{43}. Background or context: CutMix \cite{49}, Mosaic \cite{50}, CutPas \cite{56}. Multiscale: Scale and Blend \cite{57}, Simple CutPas \cite{59}. \\
  \hline
  
  Class imbalance and few images & Number of images vary between classes or some classes have only few images. 
  & Reusing instance from minority class is one strategy by instance-level operation, Simple Copy-Paste \cite{59}. Most studies attempt to generate images for the minority class: ImbCGAN \cite{84}, DAGAN \cite{86}, MFC-GAN \cite{82}, EmoGAN \cite{99}, $\delta$-encoder \cite{102}, GAN-MBD \cite{101}, SCIT \cite{100}. \\
  \hline
  
  Domain shift & Training and testing datasets represent different domains, commonly referring to styles. 
  & Changing styles for existing images is a main strategy, including S+U Learning \cite{92}, StyleAug \cite{97}, Shape bias \cite{geirhos2018imagenet}, Debiased NN \cite{103}, StyleMix \cite{104}. \\
  \hline
  \noalign{\vskip 0.2mm}
  \hline
  
  Data remembering & Larger models with many learnable parameters tend to remember specific data points, which may result in overfitting. 
  & The mechanism is increasing dataset size within or between vicinity distributions. Within version assumes label-preserving while between version changes labels, such as Mixup \cite{23}, AugMix \cite{53}, Co-Mixup \cite{kim2020co}. \\
  \hline
  
\end{tabularx}
\scriptsize
\caption{Challenges in computer vision tasks from the perspectives of datasets and deep learning models.}
\label{table:variations}
\end{table}

\begin{figure}[t]
\begin{center}
\includegraphics[width=0.5\linewidth]{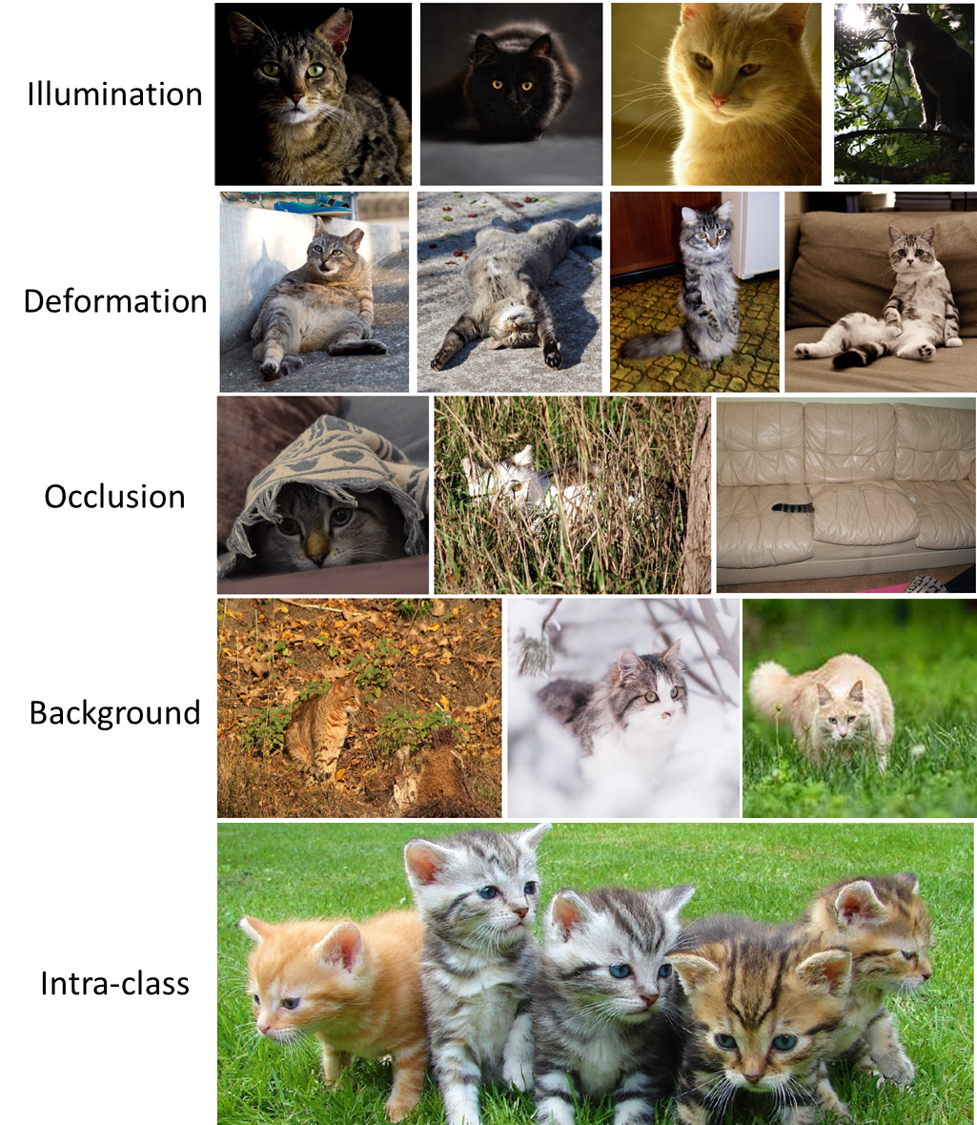}
\end{center}
\caption{Example of image variations from Class CS231n.}
\label{fig1}
\end{figure}

Table \ref{table:variations} describes the four types of challenges faced in computer vision tasks.
The first challenge is \emph{image variation}, resulting from effects such as illumination and deformation. Figure \ref{fig1} illustrates image variations\footnote{http://cs231n.stanford.edu/}. \emph{Class imbalance} is another challenge, wherein different objects are observed with different frequencies. In medical imaging, abnormal cases often occur with a low probability, which is further exacerbated by privacy. When trained with an imbalanced dataset, a model assigns a higher probability to the normal case. Besides, class imbalance becomes few images from multiple classes to one class. Furthermore, \emph{domain shift} represents a challenge where the training and testing datasets exhibit different distributions. This is exemplified by the night and day domains in the context of automatic driving. Because it is more convenient to collect images during the daytime, we may desire to train our model with a daytime dataset but evaluate it at the nighttime.

A new challenge introduced by deep learning is \emph{data remembering}. In general, a larger set of learnable parameters requires more data for training, which is referred to as structural risk \cite{18}. With an increase in parameters, a deep learning model may remember specific data points with an insufficient number of training images, which introduces a generalizability problem in the form of overfitting \cite{24}.

Fortunately, image augmentation methods can mitigate these challenges and improve model generalizability by increasing the number and variance of images in the training dataset. To utilize an image augmentation algorithm efficiently, it is crucial to understand the challenges of application and apply suitable methods. This study was conducted to provide a survey that enhances the understanding of a wide range of image augmentation algorithms.

\subsection{Vicinity distribution}
In a supervised learning paradigm, we expect to find a function $f\in \mathcal{F}$ that reflects the relationship between an input $x$ and target $y$ in a joint distribution $P(x, y)$. To learn $f$, a loss $l$ is defined to reduce the discrepancy between the prediction $f(x)$ and actual target $y$ for all examples in $P(x,y)$. We can then optimize $f$ by minimizing $l$ over $P(x, y)$, which is known as the expected risk \cite{18} and can be formulated as $R(f)=\int{l(f(x), y)dP(x,y)}$. However, $P(x,y)$ is unknown in most applications \cite{19}. Alternatively, we may use the empirical distribution $P_e(x,y)$ to approximate $P(x,y)$. In this case, the observed dataset $\mathcal{D}={(x_i,y_i )}_{i=1}^n$ is considered to be the empirical distribution, where $(x_i,y_i )$ is in $P_e(x,y)$ for a given $i$:

\begin{equation}
    P_e(x,y)=\frac{1}{n} \sum_{i=1}^n\delta((x=x_i,y=y_i)],
\end{equation}
where $\delta(x,y)$ is a Dirac mass function centered at point $(x_i,y_i)$, assuming that all masses in the probability distribution cluster around a single point \cite{21}. Another natural notion for approximating $P(x,y)$ is the vicinity distribution $P_v(x,y)$, which replaces the Dirac mass function with an estimate of the density in the vicinity of point $(x_i,y_i)$ \cite{22}:

\begin{equation}
    P_v(x,y)=\frac{1}{n}\sum_{i=1}^{n}\delta_v(x=x_i,y=y_i),
\end{equation}
where $\delta_v$ is the vicinity point set of $(x_i,y_i)$ in $\mathcal{D}$. The vicinity distribution assumes that $P(x,y)$ is smooth around any point $(x_i,y_i)$ \cite{19}. In $P_v(x,y)$, models are less prone to memorizing all data points, and thus tend to yield higher  performance in the testing process. One way to achieve vicinity distribution is to apply image augmentation, by which an original data point $(x_i,y_i)$ can be moved within its vicinity. For example, the Gaussian vicinity distribution is equivalent to the addition of Gaussian noise to an image \cite{22}.

\section{Model-free image augmentation}
Image processing methods, such as geometric transformation and pixel-level manipulation, can be leveraged for augmentation purposes \cite{9, 10}. In this study, we refer to model-free image augmentation as contrasting model-based image augmentation. The model-free approach consists of single- and multi-image branches. As suggested by the names, the former produces augmented images from a single image, whereas the latter generates output from multiple images.

\subsection{Single-image augmentation}

From the vicinity distribution, single-image augmentation (SiA) aims to fluctuate the data points in the training dataset and increase distribution density. In general, SiA leverages traditional image processing, which is simple to understand and execute. SiA methods include geometric transformations, color image processing, and intensity transformations. Geometric transformation tries to modify the spatial relationship between pixels \cite{32}, including affine transformation and elastic deformation, while color image processing aims to vary the color of an input image. In contrast, the last one is advocated to change parts of the images and has recently received more attention.

\subsubsection{Geometric transformation}
Objects in naturally captured images can appear in many variations. Geometric transformations can be employed to increase this variability. For instance, \emph{translation} provides a way to augment objects’ position. Furthermore, an image can be \emph{rotated}, changing the perspectives of objects. The angle of rotation should be carefully considered to ensure the preservation of appropriate labels. Likewise, a \emph{flip} can be executed horizontally or vertically, according to the characteristics of the training and testing datasets. For instance, the Cityscapes \cite{33} dataset can be augmented horizontally but not vertically. In addition, objects can be magnified or shrunk via \emph{scaling} to mimic multiscale variation. Finally, the \emph{elastic distortion} can alter the shape or posture of an object. Among these methods, flips have been commonly utilized throughout many studies over the last decade for various computer vision tasks, such as image classification \cite{27, 34, 35}, object detection \cite{37, ma2020mdfn}, and image translation \cite{38, XU2019570}. Two factors must be considered when using these methods: the magnitude of the operation to preserve label identity and variations in the dataset.

\subsubsection{Color image processing}
Unlike greyscale images, color images consist of three channels. Color image processing for augmentation assumes that the training and testing dataset distributions fluctuate in terms of colors, such as contrast. Although color image processing yields superior performance, it is rarely used because the color variations between the training and testing datasets are small. However, one interesting point is the use of robust features for contrast learning \cite{chen2020simple} via color image processing, which represents a case of task-agnostic learning. 

\subsubsection{Intensity transformation}

\begin{table}[h!]
\scriptsize
\begin{tabularx}{\textwidth}{ |p{0.15\textwidth}|c|X| }
  \hline
  Paper & Year & Highlight  \\
  \hline
  Hide-and-Seek \cite{40} & 2017
  & Split an image into patches that are randomly blocked.
  \\
  \hline 
  Cutout \cite{41}  & 2017
  & Apply a fixed-size mask to a random location for each image.
   \\
  \hline
  Random Erasing \cite{42} & 2020
  & Randomly select a rectangular region and displace its pixels with random values. Figure \ref{fig9}.
  \\
  \hline
  GridMask \cite{43} & 2020
  & Apply multiscale grid masks to an image to mimic occlusions. Figure \ref{fig10}.
  \\
  \hline
  
\end{tabularx}
\caption{Studies focusing upon intensity transformations. Each study is highlighted with its corresponding figure if available.}
\label{table:intensity}
\end{table}

Unlike geometric transformations and color image processing, intensity transformations entail changes at the pixel or patch levels. Random noise, such as Gaussian noise, is one of the simplest intensity transformation algorithms \cite{18}. The classical methods leverage random noise independently at the pixel level; however, the patch level has recently exhibited significant improvement for deep learning algorithms \cite{40, 41, 42, 43}. Studies pertaining to intensity transformations are listed in Table \ref{table:intensity}. The underlying concept is that the changes push the model to learn robust features by avoiding trivial solutions \cite{24}.

\begin{figure}[t]
\begin{center}
\includegraphics[width=0.6\linewidth]{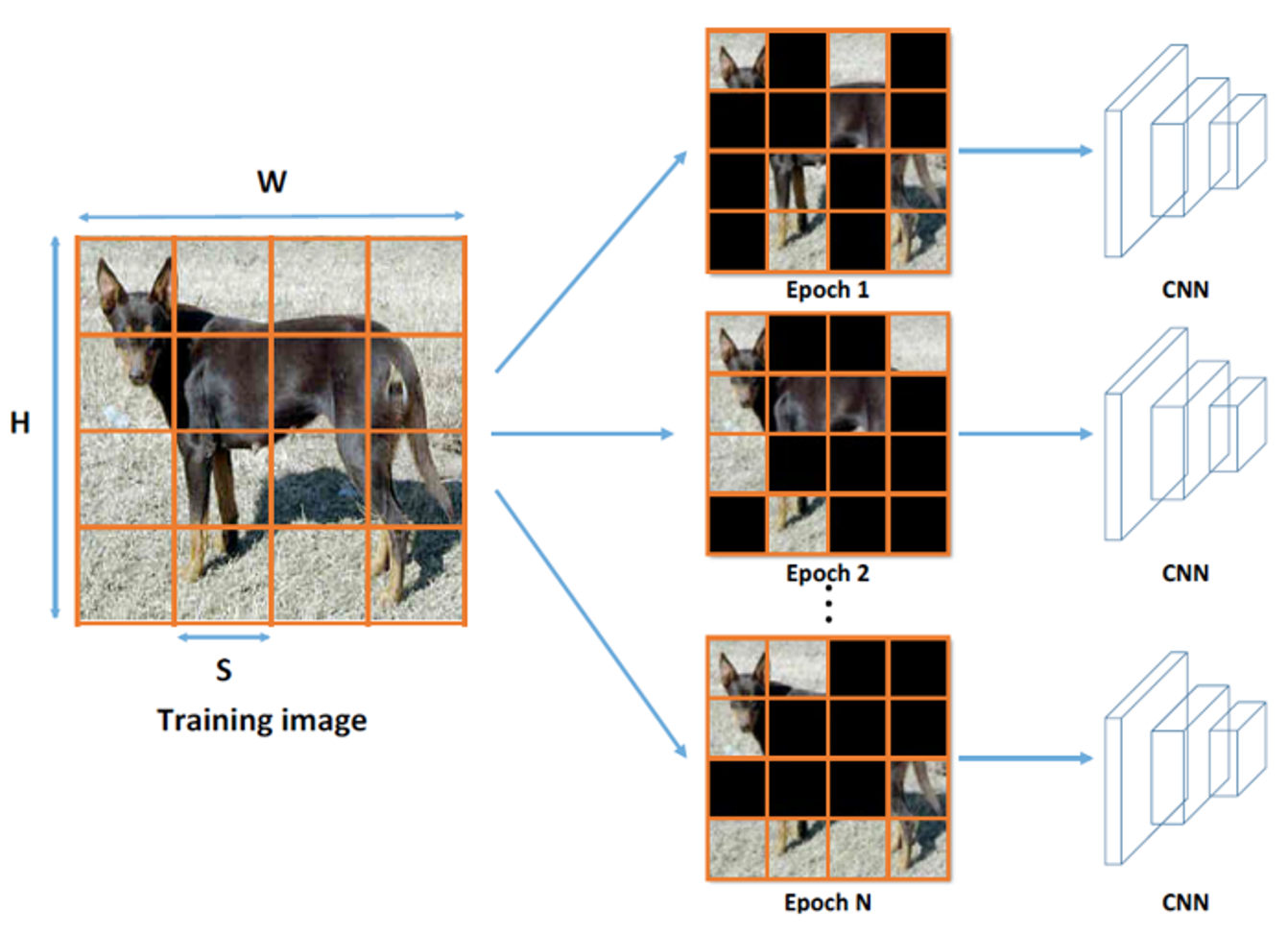}
\end{center}
\caption{Hide-and-Seek \cite{40} carries out image augmentation where one image is split into several patches, and each patch is randomly blocked with a specified probability.}
\label{fig8}
\end{figure}

\begin{figure}[t]
\begin{center}
\includegraphics[width=0.8\linewidth]{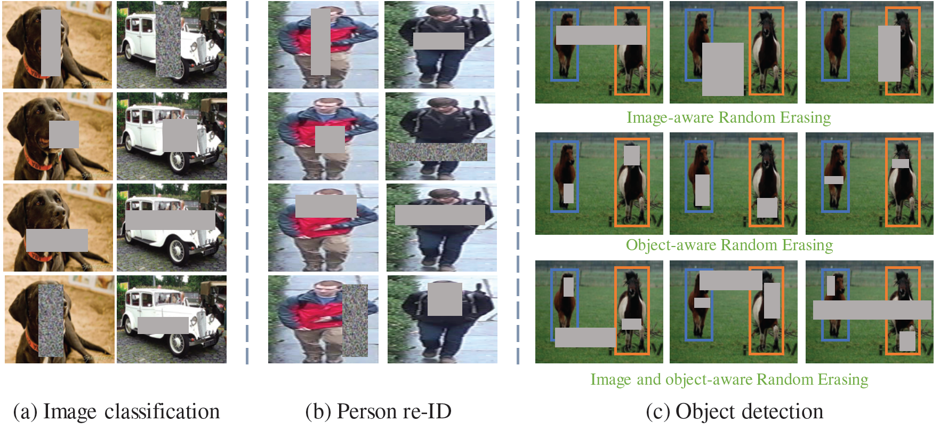}
\end{center}
\caption{Examples of Random Erasing \cite{42}.}
\label{fig9}
\end{figure}

Cutout \cite{41} randomly masks the most significant area with a finding mechanism to mimic occlusion. However, the most important aspect is cost. Hide-and-Seek \cite{40} directly blocks part of the image with the objective of obscuring the most significant area through many iterations of a random process, which is simple and fast. Figure \ref{fig8} shows that images are divided into $s\times s$ patches, and each patch is randomly blocked. One disadvantage  is that the identical size of each patch yields the same level of occlusion. To address this issue, Random Erasing \cite{42} has been employed with three random values: the size of the occluded area, height-to-width ratio, and top-left corner of the area. Figure \ref{fig9} demonstrates some examples of Random Erasing for three computer vision tasks. Additionally, this method can be leveraged in image- and object-aware conditions, thereby simplifying object detection.

\begin{figure}[t]
\begin{center}
\includegraphics[width=0.5\linewidth]{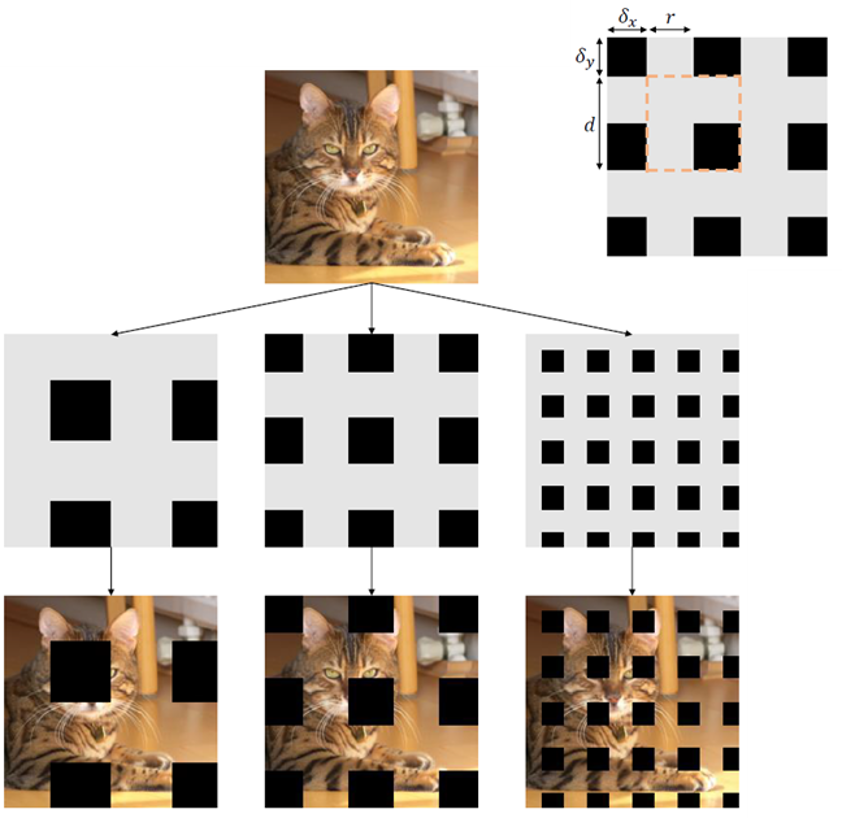}
\end{center}
\caption{GridMask \cite{43} and its setting.}
\label{fig10}
\end{figure}

GridMask aims to balance deleting and reservation, with the objective of blocking certain important areas of an object while preserving others to mimic real occlusion. To achieve this, GridMask uses a set of predefined masks, as opposed to a single mask \cite{40, 41, 42}. As illustrated in Figure \ref{fig10}, the generated mask is obtained from four values, denoting the width and height of every grid and the vertical and horizontal distance of the neighboring grid mask. By adjusting these four values, grid masks of different sizes and heigh-width ratios can be obtained. Under these conditions, GridMask achieves a better balance between deleting and reservation, and a preliminary experiment suggests that it has a lower chance of producing failure cases than Cutout \cite{41} and Hide-and-See \cite{40}.

\subsection{Multiple-image augmentation}
% Similar to single-image augmentation, multiple-image augmentation can be mathematically formulated as follows:

% \begin{equation}
%     \hat{x}=\mathcal{A}_m(x_i,x_j,...),
% \end{equation}

% where $\mathcal{A}_m$ denotes a 

Multiple-image augmentation (MiA) algorithms are executed on more than one image. These methods can further be categorized as instance- and non-instance-level. Because one image may include more than one instance, we can mask instances and use them independently. Unlike SiA, MiA requires algorithms to merge multiple input instances.

\begin{table}[h!]
\scriptsize
\begin{tabularx}{\textwidth}{ |p{0.2\textwidth}|c|X| }
  \hline
  Paper & Year & Highlight 
  \\
  \hline 
  SamplePairing \cite{47} & 2018
  & Combine two images with a single label.
  \\
  \hline
  Mixup \cite{23} & 2018
  & Linearly fuse images and their labels. Figure \ref{fig12}.
  \\
  \hline
  BC Learning \cite{48} & 2018
  & Combine two images and their labels. Treat the image as a waveform, and declare that image mixing makes sense for machines.
  \\
  \hline
  CutMix \cite{49} & 2019
  & Spatially fuse two images and linearly fuse the labels. Figure \ref{fig12}.
  \\
  \hline
  Mosaic \cite{50} & 2020
  & Spatially mix four images and their annotations, thereby enriching the context for each class.
  \\
  \hline
  AugMix \cite{53} & 2020
  & One image undergoes several basic augmentations, and the results are fused with the original image.
  \\
  \hline
  PuzzleMix \cite{51} & 2020
  & Optimize a mask for fusing two images to utilize the salient information and underlying statistics.
  \\
  \hline
  Co-Mixup \cite{kim2020co} & 2021
  & Maximize the salient signal of input images and diversity among the augmented images.
  \\
  \hline
  SuperMix \cite{52} & 2021
  & Optimize a mask for fusing two images to exploit the salient region with the Newton iterative method, 65x faster than gradient descent.
  \\
  \hline
  GridMix \cite{54} & 2021
  & Split two images into patches, spatially fuse the patches, and linearly merge the annotation.
  \\
  \hline
  \noalign{\vskip 0.2mm}
  \hline
  Cut, Paste and Learn \cite{56} & 2017
  & Cut object instances and paste them onto random backgrounds. Figure \ref{fig16}.
  \\
  \hline
  Scale and Blend \cite{57} & 2017
  & Cut and scale object instances, and blend them in meaningful locations.
  \\
  \hline
  Context DA \cite{58} & 2018 
  & Combine object instances using context guidance to obtain meaningful images.
   \\
  \hline
  Simple Copy-Paste \cite{59} & 2021 
  & Randomly paste object instances to images with large-scale jittering. 
   \\
  \hline
  Continuous Copy-Paste \cite{60} & 2021 
  & Deploy Cut, Paste and Learn to videos. 
  \\
  \hline
  
\end{tabularx}
\caption{Studies related to multiple-image augmentation, divided into non-instance- (up) and instance-level (bottom).}
\label{table:image-mix}
\end{table}

\subsubsection{Non-instance-level}

\begin{figure}[t]
\begin{center}
\includegraphics[width=0.6\linewidth]{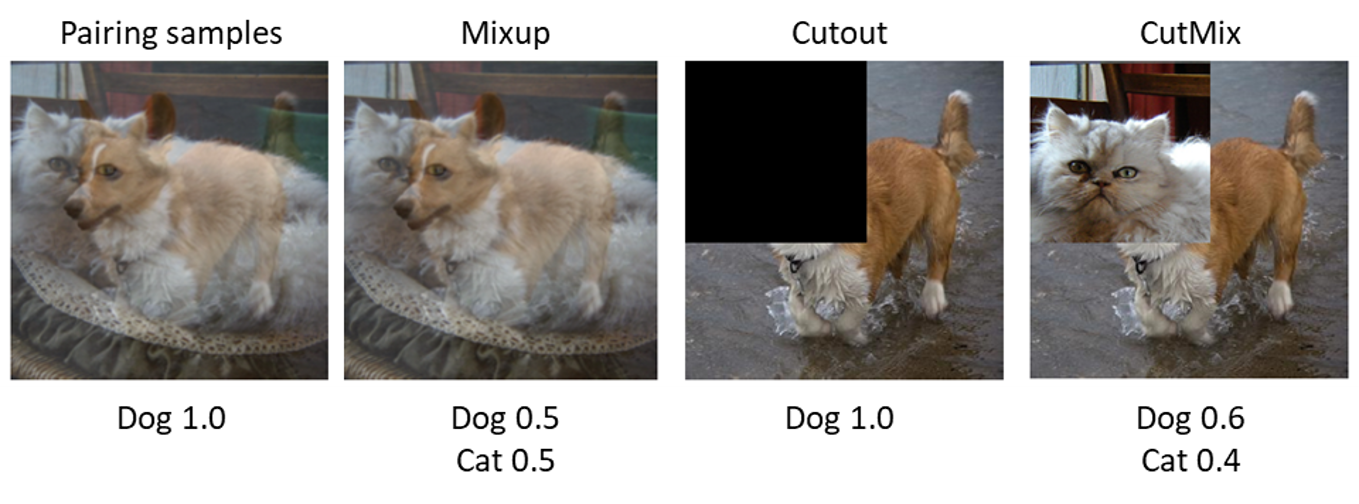}
\end{center}
\caption{Comparison of non-instance-level multiple-image algorithms \cite{49}.}
\label{fig12}
\end{figure}

In the context of MiA algorithms, the non-instance-level approach adopts and fuses the images. Studies pertaining to this concept are listed in Table \ref{table:image-mix}. One of the simplest methods is to compute the average value of each pixel. In Pairing Samples \cite{47}, two images are fused to produce an augmented image with a label from one source image. This assumption is generalized in Mixup \cite{23}, where the labels are also fused. Figure \ref{fig12} illustrates the difference between Pairing Samples and Mixup. Mathematically, $\tilde{x}=\lambda x_i + (1- \lambda) x_j$ and $\tilde{y}=\lambda y_i + (1- \lambda) y_j$,
where $x_i$ and $x_j$ are two images, $y_i$ and $y_j$ are the corresponding one-hot labels, and $\tilde{x}$ and $\tilde{y}$  denote the generated image and label, respectively. By adjusting $0 \leq \lambda \leq 1$, many images with different labels can be created, thereby smoothing out the gap between the two labels in the augmented images. Although Pairing Samples and Mixup produce satisfactory results, the fused images are not reasonable for humans. Accordingly, these fused images have been declared to make sense for machines from the perspective of a waveform \cite{48}. In addition, vicinity distribution can also be utilized to understand this situation. To be more specific, changing image variations yet maintaining the label can be regarded a deviation in the vicinity distribution space of a specific label, whereas image fusion can be considered as an interpolation between the vicinity distribution of two labels \cite{23}.

% \begin{figure}[t]
% \begin{center}
% \includegraphics[width=0.7\linewidth]{14.png}
% \end{center}
% \caption{Examples of Mosaic image augmentation \cite{50}.}
% \label{fig14}
% \end{figure}

In contrast to BC Learning \cite{48}, CutMix \cite{49} spatially merges images to obtain results that are interpretable by humans. The last picture in Figure \ref{fig12} illustrates the method’s underlying strategy, wherein the merged image consists of two source images spatially, and its label is obtained from the ratio of certain pixels between two images. Although multiple-image augmentation generally utilizes two images, more images can be used. For example, Mosaic \cite{50} employs four images wherein the number of objects in one image is increased, thus significantly reducing the need for a large mini-batch size for dense prediction. AugMix \cite{53} randomly applies basic multiple methods of image augmentation, and the results are adopted to merge with the original image. 

Non-instance-level image augmentation has extensions similar to those of intensity transformations. To account for the most important area, PuzzleMix \cite{51} discriminates the foreground from the background, and mixes important information within the foreground. Further, salient areas from multiple input images are maximized to synthesize each augmented image \cite{kim2020co}, simultaneously maximizing the diversity among the augmented images. To quickly locate dominant regions, SuperMix \cite{52} employs a variant of the Newton iterative method. As in Hide-and-Seek \cite{40}, GridMix \cite{54} divides images into fixed-size grids, and each patch of the output image is randomly taken from the corresponding patches of two input images. Through this analysis, we believe that GridMask \cite{55} can be adapted to fuse image pairs with changeable sizes.

\subsubsection{Instance-level}

\begin{figure}
\begin{center}
\includegraphics[width=0.5\linewidth]{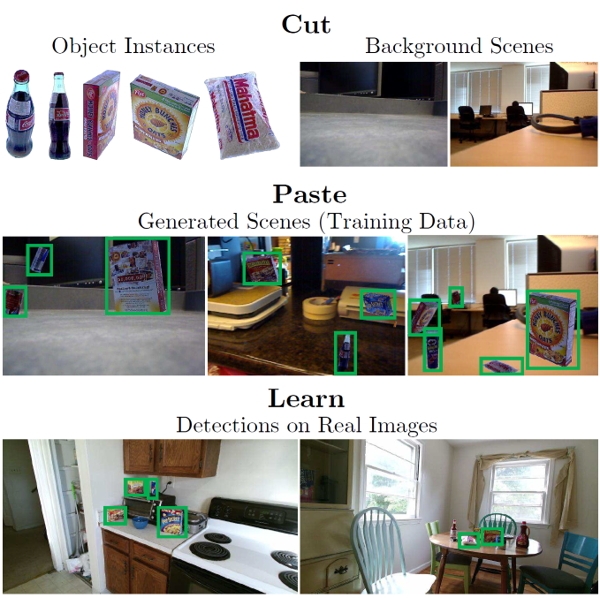}
\end{center}
\caption{Cut, Paste and Learn in training and testing process \cite{56}.}
\label{fig16}
\end{figure}

Whereas the non-instance-level approach employs images directly, the instance-level approach leverages instances masked from images. Related studies are listed in the second part of Table \ref{table:image-mix}. The instance-level approach comprises two main steps. As shown in Figure \ref{fig16}, the first step involves cutting instances from source images given a semantic mask, and obtaining clean background senses. Next, the obtained instances and background are merged. Cut, Paste and Learn \cite{56} is an early instance-level method, wherein local artifacts are noticed after pasting instances to the background. Because local region-based features are important for object detection, various blending modes are employed to reduce local artifacts. With the exception of boundaries, the instance scale and position are not trivial, as objects may be multiscale and recognizable with the help of their contexts, as addressed in \cite{57}. 

% \begin{figure}
% \begin{center}
% \includegraphics[width=0.8\linewidth]{17.png}
% \end{center}
% \caption{Overview to blend instance and background \cite{57}, in which two issues in the blending process are noticed, where to put the instance and the scale of the instance with the help of depth image (e). (b) and (f) is the semantic segmentation from a trained model and the plane extraction from the depth image (e). To put the instance, tables and counters are predicted from segmentation prediction (b), and the support surface (g) is estimated from plane extraction. After choosing the place to put as shown in (d), the desired instance can be scaled on a different scale and then is put in the chosen places to form the augmented image (f).}
% \label{fig17}
% \end{figure}

% As shown in Figure \ref{fig17}, tables and counters are predicted with a trained segmentation model, and the support surface is estimated from the depth image. The two components are utilized to decide where to place the desired instance, as (d) in the figure. Moreover, the scale of the desired instance is randomly changed, and the rescaled instance is placed in the supposed position. The position and scale issue was verified to be promising in indoor scenes in \cite{57}, and has also been proven for outdoor scenes in \cite{58}, where context information is utilized to situate the desired instance.

Interestingly, instance-level image augmentation can mitigate the challenges posed by class imbalance. By repurposing rare instances, the number of images in the corresponding class increases. Simple Copy-Paste \cite{59} indicates that the instance level enables strong image augmentation methods, for instance, segmentation. While it is based on Copy, Paste and Learn, Simple Copy-Paste differs in two characteristics. First, the background image is randomly selected from the dataset, and subjected to random scale jittering and horizontal flipping. Second, large-scale jittering is leveraged to obtain more significant performance. The copy-paste concept has also been utilized for time-series tasks \cite{60} such as tracking.

\section{Model-based image augmentation}

% \begin{figure}[t]
% \begin{center}
% \includegraphics[width=0.5\linewidth]{20.png}
% \end{center}
% \caption{The difference between model-free and model-based image augmentation. Model-based image augmentation requires training a model which is utilized to generate augmented images given extra conditions.}
% \label{fig20}
% \end{figure}

A model must be pre-trained in model-based image augmentation to generate augmented images. The present study classifies this process among three categories, according to the conditions to generate images: unconditional, label-conditional, and image-conditional. Table \ref{tab:my_label} provides information regarding appropriate studies.
% There are essential questions in model-based image augmentation. For example, what types of models should be used and how should they be trained? However, we focused on explaining the reason for performing image augmentation.

\begin{table}[h!]
    \centering
    \scriptsize
    \begin{tabularx}{\textwidth}{ |r|c|X| }
         \hline
         Paper & Year & Highlight  \\
         
         \hline
         BDA \cite{79} & 2017 
         & Use CGAN to generate images optimized by a Monte Carlo EM algorithm. Figure \ref{fig21}.
         \\
         
          \hline
         ImbCGAN \cite{84} & 2018 
         & Deploy CGAN as image augmentation for unbalanced classes.
         \\
         
         \hline
         BAGAN \cite{83} & 2018
         & Train an auto-encoder to initialize generator.
          \\
         
         \hline
         DAGAN \cite{86} & 2018
         & Image is taken as the class condition for generator and discriminator. Figure \ref{fig22}.
         \\
         
        \hline
         MFC-GAN \cite{82} & 2019
         & Use multiple fake classes to obtain a fine-grained image for minority class.
          \\
         
         \hline
         IDA-GAN\cite{85} & 2021
         & Train a variational auto-encoder and CGAN simultaneously.
         \\
         
         \hline
         \noalign{\vskip 0.2 mm}
         
         \hline
         S$+$U Learning \cite{92} & 2017
         & Translate synthetic images from a graphic model to realistic images using CGAN.
         \\
         
         \hline
         AugGAN \cite{91} & 2018
         & Aim to semantically preserve object when changing its style.
         \\
         
         \hline
         Plant-CGAN \cite{93} & 2018
         & Translate semantic instance layout to real images using CGAN.
         \\
         
         \hline
         StyleAug \cite{97} & 2019
         & Change image style via style transfer.
          \\
         
         \hline
         Shape bias \cite{geirhos2018imagenet} & 2019
         & Transfer image style from painted images to mitigate texture bias of CNN.
         \\
         
         \hline
         \noalign{\vskip 0.2mm}
         
         \hline
         EmoGAN \cite{99} & 2018
         & Translate a neutral  face with another emotion.
         \\
         
         \hline
         $\delta$-encoder \cite{102} & 2018
         & Image is taken as a class condition to generate images for new or infrequent class.
         \\
         
         \hline
         Debiased NN \cite{103} & 2021
         & Merge style and content via style transfer and appropriate labels. Figure \ref{fig27}.
         \\
         
         \hline
         StyleMix \cite{104} & 2021
         & Merge two images with style, content, and labels. Figure \ref{fig28}.
         \\
         
         \hline
         GAN-MBD \cite{101} & 2021
         & Translate an image from one class to another while preserving semantics via multi-branch discriminator. Figure \ref{fig25}.
         \\
         
         \hline
         SCIT \cite{100} & 2022
         & Translate healthy leaves to abnormal one while retaining its style.
         \\
         
         \hline
    \end{tabularx}
    \caption{Studies relating to model-based image augmentation, label-conditional (top), label-preserving image-conditional (middle), and label-changing image-conditional (bottom).}
    \label{tab:my_label}
\end{table}

\subsection{Unconditional image generation}
An image synthesis model benefits image augmentation, which enables it to produce new images. Theoretically, the distribution of generated images is similar to that in the original dataset for a generative adversarial network (GAN) model after training \cite{68}. However, the generated images are not the same as the original images and can be considered as points located in the vicinity distribution. In DCGAN \cite{75}, two random noises or latent vectors can be interpolated to generate intermediate images, which can be regarded as fluctuations between two original data points. Generally, a generative model with noise as input is deemed an unconditional model, and the corresponding image generation process is considered unconditional image generation. If the datasets encompass a single class, as in the case of medical images with one abnormal class \cite{76}, an unconditional image generation model can be directly applied to perform augmentation. Furthermore, a specific unconditional model can be leveraged for an individual class in the presence of multiple classes \cite{77}, \cite{78}.

\subsection{Label-conditional image generation}
Although unconditional image generation has potential, the shared information of different classes cannot be utilized. In contrast, label-conditional image generation is expected to leverage the shared information and learn variations for minority classes using majority-class data. Label-conditional image generation requires one specific label as an extra input, and the generated image should align with the label condition.

\begin{figure}[t]
\begin{center}
\includegraphics[width=0.9\linewidth]{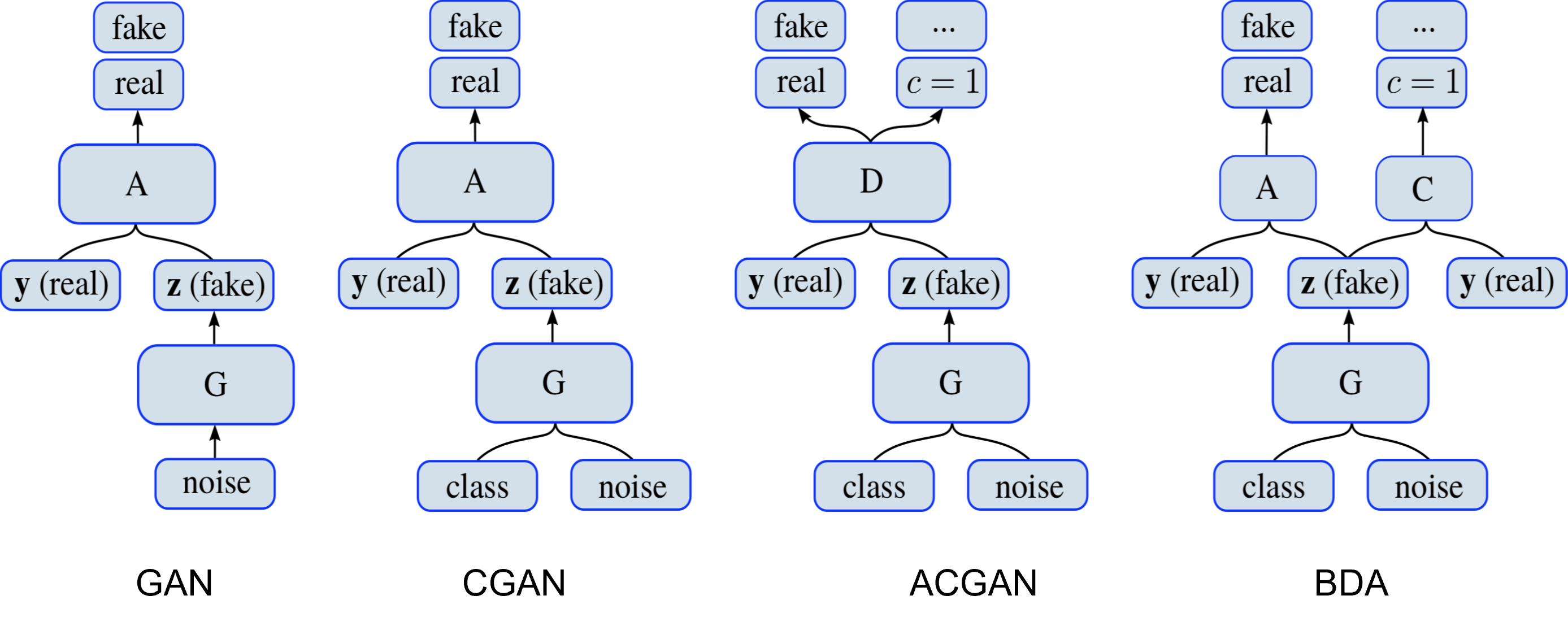}
\end{center}
\caption{GAN and variants of label-conditional GANs \cite{79}. G: generator, A: authenticator, C: classifier, D: discriminator.}
\label{fig21}
\end{figure}

The primary issue in label-conditional image generation is the use of label conditions. CGAN \cite{80} uses  the label for a generator, whereas the authenticator does not use the label. Consequently, the generator tends to ignore label information, as the authenticator cannot provide feedback regarding the condition. ACGAN \cite{81} introduces an auxiliary classifier in the discriminator, which encourages the generator to produce images aligned with label conditions. With a more complex classifier, BDA \cite{79} separates the classifier from the discriminator. Figure \ref{fig21} illustrates the differences between BDA and other label-conditional algorithms. In addition, MFC-GAN \cite{82} adopts multiple fake classes in the classification loss to stabilize the training.

One of the main applications of label-conditional image generation is the class imbalance \cite{82} \cite{84} \cite{85}. The generative model is expected to learn useful features from the majority class, and use them to generate images for the minority classes. The generated images are used to rebalance the original training dataset. However, it may be challenging to train a GAN model with an unbalanced dataset, as the majority class dominates the discriminator loss and the generator tends to produce images within the majority class. To address this challenge, a pretrained autoencoder with reconstruction loss has been employed to initialize a generator \cite{83} \cite{85}.

\begin{figure}[t]
\begin{center}
\includegraphics[width=0.85\linewidth]{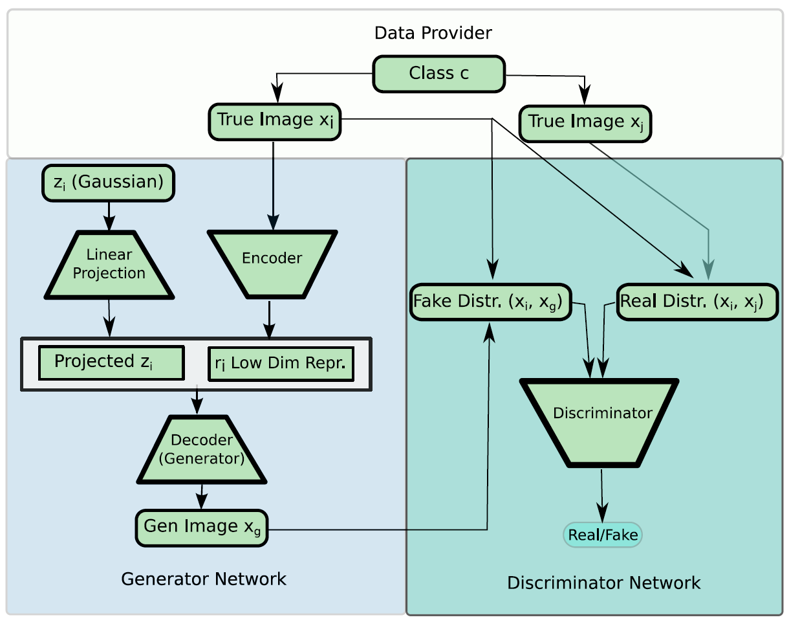}
\end{center}
\caption{Flowchart of DAGAN \cite{86}, where label information is obtained from an image via an encoder, rather than a label.}
\label{fig22}
\end{figure}

Although various discriminators and classifiers may be employed, the aforementioned algorithms utilize the class condition on a one-hot label. One resulting limitation is that the trained model can generate only known-class images. To overcome this limitation, DAGAN \cite{86} utilizes an image encoder to extract the class, so that the generated image is assumed to have the same class as the original image. Figure \ref{fig22} illustrates the DAGAN algorithm.

\subsection{Image-conditional image generation}

In image generation, images can be employed as conditions, known as image translation.  Generally, an image consists of content and style \cite{87, 88}. Content refers to class-dependent attributes, such as dogs and cats, whereas style denotes class-independent elements, such as color and illumination. Image-conditional image generation can be subcategorized into two types: label-preserving and label-changing. The former requires content to be retained, whereas the latter requires content to be changed.

\subsubsection{Label-preserving image generation}

Label-preserving assumes that the label of a generated image is the same as that of the input image. One active field to deploy this approach is the domain shift, where the style of the source domain is different from that of the target domain. To address this challenge, original images can be translated from the source domain to the target domain. To preserve the object during image translation, AugGAN employs a segmentation module that extracts context-aware features to share parameters with a generator \cite{91}. For practical applications, synthetic images generated by a graphical model are translated into natural images \cite{92}, and the leaf layout is translated as a real leaf image \cite{93}. In addition, image translation can be utilized for semantic segmentation with a domain shift \cite{li2020simplified}. Furthermore, label-preserving can be leveraged to improve the robustness of a trained model. Inspired by the observation that CNNs exhibit bias on texture toward shape, original images are translated to have different textures, which allows the CNN to allocate more attention to shape \cite{geirhos2018imagenet}.

It is often challenging to obtain the desired style during the image generation process. Most algorithms utilize an encoder to extract style from an image, as in the case of DRIT++ \cite{90} and SPADE \cite{39}. This approach to image translation can be regarded as image fusion. In contrast, Jackson et al. \cite{97} proposed style augmentation, where the style is generated from a multivariate normal distribution. Another challenge is that the \emph{one} model can be adopted to generate images for \emph{multiple} domains with fewer trained images. To address this, MetalGAN leverages domain loss and meta-learning strategies \cite{98}.

\subsubsection{Label-changing image generation}

\begin{figure}[t]
\begin{center}
\includegraphics[width=0.5\linewidth]{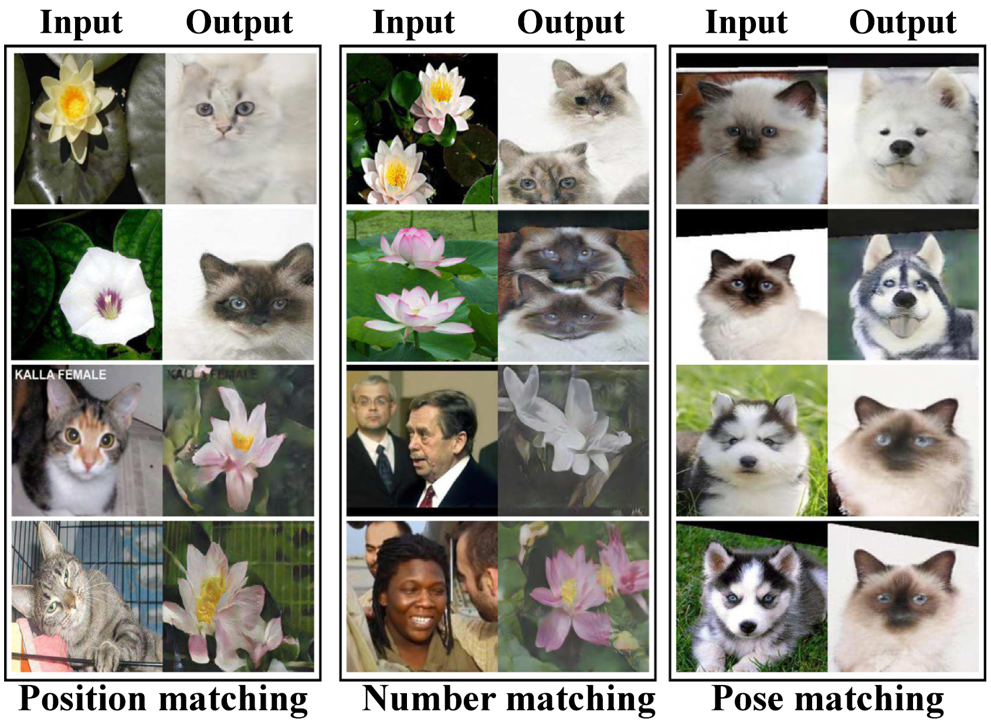}
\end{center}
\caption{Semantic level matching by GAN-MBD \cite{101} for label-changing image augmentation, including position, number, and pose.}
\label{fig25}
\end{figure}

In contrast to label-preserving, label-changing changes the label-dependent. For example, a neutral face can be transformed into a different emotion \cite{99}. Although the generated images have poor fidelity, the approach improves the classification of emotions. In addition to changing label dependence, the preservation of label independence has recently received attention as a way to improve variability within the target class, thereby mitigating class imbalance. To take variation from one to another class, a style loss is leveraged to retain the style when translating an image \cite{100}. Similarly, a multi-branch discriminator with fewer channels is introduced to achieve semantic consistency such as the number of objects \cite{101}. Figure \ref{fig25} shows several satisfactory translated images. To address severe class imbalance, a $\delta$-encoder has been proposed to extract label-independent features from one label to another \cite{102}. As in the case of DAGAN \cite{86}, class information is provided by an image. The $\delta$-encoder and decoder aim to reconstruct the given image in the training phase, whereas the decoder is provided a new label image and required to generate the same label in the testing phase.

\begin{figure}[t]
\begin{center}
\includegraphics[width=0.6\linewidth]{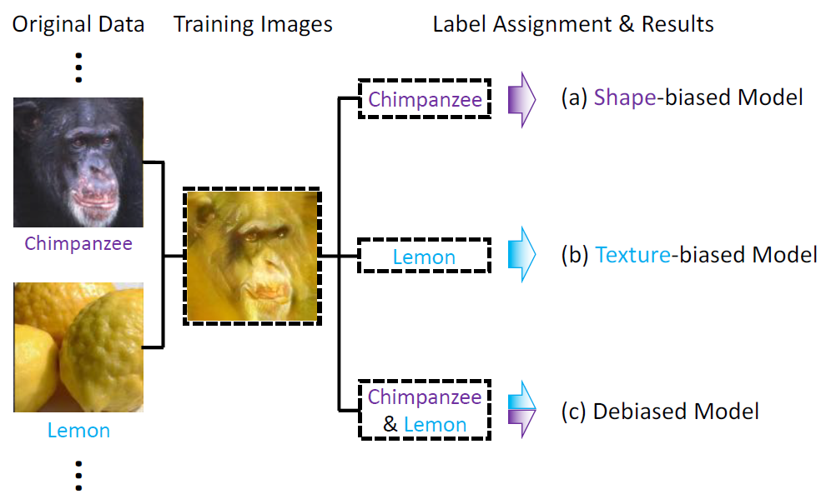}
\end{center}
\caption{Label assignment for the biased and unbiased model with respect to shape and texture \cite{103}.}
\label{fig27}
\end{figure}

\begin{figure}[t]
\begin{center}
\includegraphics[width=0.5\linewidth]{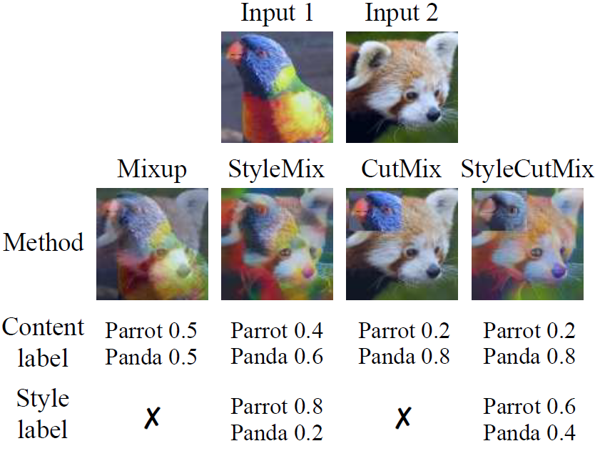}
\end{center}
\caption{Examples of label assignment with different algorithms \cite{104}.}
\label{fig28}
\end{figure}

Compared to label-preserving, label-changing yields more significant improvements in model robustness by changing the label and style simultaneously.
% Although humans can recognize objects with different styles, machines tend to be biased towards style or class dependence \cite{103}. 
As illustrated in Figure \ref{fig27}, traditional image augmentation does not change the label after altering the color of the chimpanzee to that of a lemon, which incurs shape bias. By contrast, when a texture-biased model is trained, the translated image is labeled as a lemon. To balance the bias, the translated image by style transfer is taken with two labels \cite{103} -- chimpanzee and lemon -- which eliminates bias. Inspired by Mixup \cite{23}, Hong et al. developed StyleMix \cite{104}, which merges the two inputs to obtain content and style labels, as shown in Figure \ref{fig28}. These labels are then fused to obtain the final label for the generated images.

\section{Optimizing policy-based image augmentation}

All algorithms mentioned in the previous two sections represent specific schemes, wherein domain knowledge is required to achieve better performance. In general, individual operations with the desired magnitude are utilized to perform image augmentation for specific datasets according to their characteristics. However, hyperparameter optimization is challenging and time-consuming. One way to mitigate this is to design algorithms that determine optimal augmentation strategies. These algorithms, termed policy-based optimization, encompass two categories: reinforcement learning-based, and adversarial learning-based. The former category employs reinforcement learning (RL) to determine the optimal strategy, whereas the latter category adopts augmented operations and their magnitudes that generates a large training loss and small validation loss. As generative adversarial networks (GANs) can be utilized for both model-based and optimizing policy-based image augmentation, the objective to adopt GANs is the primary difference. Model-based category aims to \emph{directly} generate images, instead of other goals such as finding optimal transformations \cite{116}. Studies pertaining to policy-based optimization are listed in Table \ref{table:optim}.

\begin{table}[h!]
    \centering
    \scriptsize
    \begin{tabularx}{\textwidth}{|p{0.2\textwidth}|c|X|}
        \hline
        Paper & Year & Highlight  \\
        
        \hline
        AutoAugment \cite{105} & 2019
        & Use reinforcement learning to determine the optimal augmentation strategies. Figure \ref{fig29}.
        \\
        
        \hline
        Fast AA \cite{107} & 2019
        & Use efficient density matching for augmentation policy search.
        \\
        
        \hline
        PBA \cite{109} & 2019
        & Adopt non-stationary augmentation policy schedules via population-based training.
        \\
        
        \hline
        Faster AA \cite{108} & 2019
        & Use a differentiable policy search pipeline via approximate gradients.
        \\
        
        \hline
        RandAugment \cite{106} & 2020
        & Reduce the search space of AutoAug via probability adjustment.
        \\
        
        \hline
        MADAO \cite{110} & 2020
        & Train task model and optimize the search space simultaneously by implicit gradient with Neumann series approximation.
        \\
        
        \hline
        LDA \cite{114} & 2020
        & Take policy search as a discrete optimization for object detection.
        \\
        
        \hline
        LSSP \cite{113} & 2021 
        & Learn a sample-specific policy for sequential image augmentation.
        \\
        
        \hline
        \noalign{\vskip 0.2mm}
        \hline
    
        \hline
        ADA \cite{116} & 2016
        & Seek a small transformation that yields maximal classification loss on the transformed sample.
        \\
        
        \hline
        CDST-DA \cite{117} & 2017
        & Optimize a generative sequence using GAN in which the transformed image is pushed to be within the same class distribution.
        \\
        
        \hline
        AdaTransform \cite{115} & 2019
        & Use a competitive task to obtain augmented images with a high task loss in the training stage, and a cooperative task to obtain augmented images with a low task loss in the testing stage. Figure \ref{fig31}.
        \\
        
        \hline
        Adversarial AA \cite{111} & 2020
        & Optimize a policy to increase task loss while allowing task model to minimize the loss.
        \\
        
        \hline
        IF-DA \cite{119} & 2020
        & Use influence function to predict how validation loss is affected by image augmentation, and minimize the approximated validation loss.
        \\
        
        \hline
        SPA \cite{118} & 2021
        & Select suitable samples to perform image augmentation.
        \\
        
        \hline
        
    \end{tabularx}
    \caption{Studies relating to optimizing policy of image augmentation. The upper and the bottom suggest reinforcement learning- and adversarial learning-based image augmentation.}
    \label{table:optim}
\end{table}

\subsection{Reinforcement learning-based}

\begin{figure}[t]
\begin{center}
\includegraphics[width=0.6\linewidth]{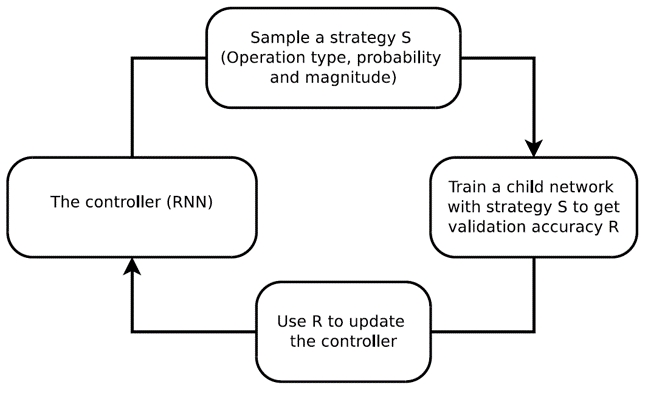}
\end{center}
\caption{Overview of AutoAugment \cite{105}, a reinforcement learning-based image augmentation method.}
\label{fig29}
\end{figure}

AutoAugment \cite{105} is a seminal approach that employs reinforcement learning. As shown in Figure \ref{fig29}, iterative steps are used to find the optimal policy. The controller samples a strategy from a search space with the operation type and its corresponding probability and magnitude, and a task network subsequently obtains the validation accuracy as feedback to update the controller. Because the search space is very large, lighter child networks are leveraged. After training, the controller is used to train the original task model and can be finetuned in other datasets. 

Although AutoAugment achieves satisfactory classification performance across several datasets, it requires a long training time. To address this issue,  several studies have been conducted from different perspectives. For instance, RandAugment \cite{106} replaces several probabilities in AutoAugment with a uniform probability. Conversely, Fast AA \cite{107} and Faster AA \cite{108} leverage density matching, aligning the densities of the training and augmented training datasets, instead of Proximal Policy Optimization \cite{schulman2017proximal}, to optimize the controller in AutoAugment. Furthermore, PBA \cite{109} attempts to learn a policy schedule from population-based training, rather than a single policy.

Except for the long training phase, AutoAugment utilizes child models, by which the learned policy may not be optimal for the final task model. To address this issue, Hataya et al. \cite{110} trained the target model and image augmentation policy simultaneously using the same differentiable image augmentation pipeline in Faster AA. In contrast, Adversarial AA \cite{111} leverages adversarial loss simultaneously with reinforcement learning.

One limitation of the algorithms mentioned above is that the learned image augmentation policy is at the dataset level. Conversely, class- and sample-level image augmentation methods were considered in \cite{112} and \cite{113}, respectively, wherein each class or sample utilizes a specific policy. Furthermore, instance-level image augmentation was considered in \cite{114} for object detection, where operations were performed only inside the bounding box.

\subsection{Adversarial learning-based}

The primary objective of image augmentation is to train a task model with a training dataset to achieve sufficient generalizability on a testing dataset. One assumption is that hard samples are more useful, and the input images that cause a larger training loss are considered hard samples. Adversarial learning-based image augmentation aims to learn an image augmentation policy to generate hard samples based on the original training samples.

An early method \cite{116} attempts to find a small transformation that maximizes training loss on the augmented samples, wherein learning optimization finds an optimal magnitude given an operation. One of the main limitations is the label-preserving assumption that the augmented image retains the same label as the original image. To meet this assumption, a common strategy is to design the type of operation and range of corresponding magnitude using human knowledge. To weaken this assumption, Ratner et al. \cite{117} introduced generative adversarial loss to learn a transformation sequence in which the discriminator pushes the generated images to one of the original classes, instead of an unseen or null class.

\begin{figure}[h!]
\begin{center}
\includegraphics[width=1.0\linewidth]{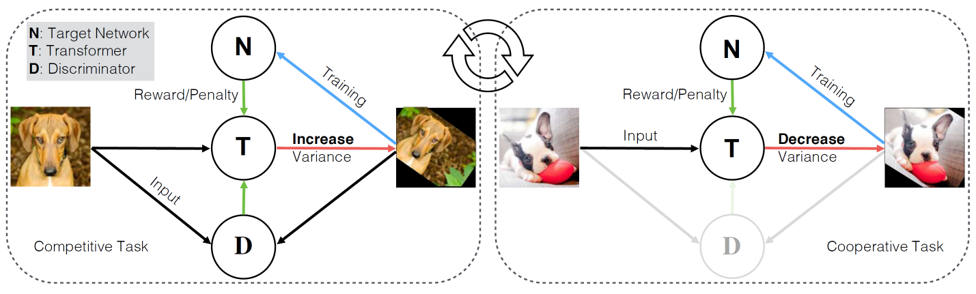}
\end{center}
\caption{Overview of AdaTransform \cite{115}. AdaTransform encompasses two tasks -- competitive training and cooperative testing -- and three components: transformer $T$, discriminator $D$, and target network $N$. The transformer increases the variance of training data by competing with both $D$ and $N$. It also cooperates with $N$ in the testing phase to reduce data variance.}
\label{fig31}
\end{figure}

Interestingly, SPA \cite{118} attempts to select suitable samples, and image augmentation is leveraged only on those samples in which the augmented image incurs a larger training loss than the original image. Although SPA trains the image augmentation policy and task model simultaneously at the sample level, the impact of the learned policy in the validation dataset is unknown. To address this challenge, an influence function was adopted for approximating the change in validation loss without actually comparing performance \cite{119}. Another interesting concept is the use of image augmentation in the \emph{testing stage}. To achieve this, AdaTransform \cite{115} learns two tasks -- competitive and cooperative -- as illustrated in Figure \ref{fig31}. In a competitive task, the transformer learns to increase the input variance by increasing the loss of the target network, while the discriminator attempts to push the augmented image realistically. Conversely, the transformer learns to decrease the variance of the augmented image in the cooperative task by reducing the loss of the target network. After training, the transformer is utilized to reduce the variance of the input image, thereby simplifying the testing process.

\section{Discussions}

In this section, the usage of the mentioned strategies to perform image augmentation are first discussed. Several future directions are then illustrated. Furthermore, three related topics are discussed: understanding image augmentation from theory perspective, adopting image augmentation with other strategy, and augmenting features instead of images.

\textbf{Current situation.}
Datasets are assumed to be essential to obtain satisfactory performance. One way to generate an appropriate dataset is through image augmentation algorithms, which have demonstrated impressive results across multiple datasets and heterogeneous models. For instance, Mixup \cite{23} increases the validation accuracy in ImageNet-2012 by 1.5 and 1.2 percent with ResNet-50 and ResNet-101. Non-trivially, GAN-MBD \cite{101} achieves 84.28 classification accuracy with an unbalance dataset setting in 102Flowers, 33.11, 31.44, and 14.05 higher than non-image augmentation, geometrical transformation, and focal loss, respectively. Currently, mode-free and optimizing policies are widely leveraged, whereas the mode-based approach is an active research topic for specific challenges, such as class imbalance and domain adaptation. In addition, although most algorithms are label-preserving, label-changing algorithms have recently received attention.

\textbf{Future direction.} Although many image augmentation algorithms exist, developing novel algorithms remains crucial to improve the performance of deep learning. We argue that recognizing new challenges or variations may inspire novel methods if they can be mimicked using image augmentation. Further, most algorithms of image augmentation are designed for classification and hence extending them to other applications is one of the most applicable directions by incorporating application-based knowledge, such as time-series in video \cite{60}. Another interesting direction is distinguishing specific applications from general computer vision tasks such as ImageNet \cite{69} and COCO \cite{70} and then finding new motivations to design image augmentation. For example, most variations in plant healthy and diseased leaves are shared and thus can be converted from one to another \cite{100}. Finally, considering image augmentation from a systematic perspective is appealing. For example, the effects of image augmentation schedules on optimization such as learning rate and batch size, are analyzed in \cite{hanin2021data}.

\textbf{Understanding image augmentation.} This study was conducted to understand the objectives of image augmentation in the context of deep learning, from the perspectives of challenges and vicinity distribution. Although it was also verified that image augmentation is similar to regularization \cite{22}, most of the evidences are empirically from experiments. Understanding them in theory is therefore appealing. Recently, kernel theory \cite{dao2019kernel} and group theory \cite{chen2020group} have been used to analyze the effects of image augmentation. In addition, the improvement yielded by image augmentation in the context of model generalizability has been quantified using affinity and diversity \cite{gontijo2020tradeoffs}.

\textbf{New strategy to leverage image augmentation.} 
Although image augmentation is commonly used in a supervised manner, this must not necessarily be the case. First, a pretext task can be created via image augmentation, such as predicting the degrees of rotation \cite{komodakis2018unsupervised} and relative positions of image patches \cite{doersch2015unsupervised}. Second, image augmentation can be leveraged to generate positive samples for contrast learning under the assumption that an augmented image is similar to the corresponding original image \cite{ye2019unsupervised, grill2020bootstrap, caron2021emerging}. Furthermore, semi-supervised learning benefits from image augmentation \cite{22, berthelot2019remixmatch, xie2020unsupervised}.

\textbf{Feature augmentation} attempts to perform augmentation in feature space instead of image space in image augmentation, and thus reduces the computation cost but without visual evidences. A feature space generally has dense information in semantic level than an image space. Consequently, operation in feature space is more efficient \cite{wang2021regularizing}, such as domain knowledge \cite{wang2020data}. Simultaneously, we believe that most of the techniques in image augmentation can be extended to feature augmentation, such as Manifold Mixup \cite{verma2019manifold} from Mixup \cite{23} and occluded feature \cite{cen2021deep}.

\section{Conclusion}
This study surveyed a wide range of image augmentation algorithms with a novel taxonomy encompassing three categories: model-free, model-based, and optimizing policy-based. To understand the objectives of image augmentation, we analyzed the challenges of deploying a deep learning model for computer vision tasks, and adopted the concept of vicinity distribution. We found that image augmentation significantly improves task performance, and many algorithms have been designed for specific challenges, such as intensity transformations for occlusion, and model-based algorithms for class imbalance and domain shift. Based on this analysis, we argue that novel methods can be inspired by new challenges. Conversely, appropriate methods can be selected after recognizing the challenges posed by a dataset. Furthermore, we discussed the current situation and possible directions of image augmentation with three relevant interesting topics. We hope that our study will provide an enhanced understanding of image augmentation and encourage the community to prioritize dataset characteristics.

\section*{Acknowledgment}
This research was partly supported by the Basic Science Research Program through the National Research Foundation of Korea (NRF) funded by the Ministry of Education (No.2019R1A6A1A09031717), supported by the National Research Foundation of Korea (NRF) grant funded by the Ministry of Science and ICT (MSIT) (No. 2020R1A2C2013060), and supported by the Korea Institute of Planning and Evaluation for Technology in Food, Agriculture, and Forestry (IPET) and Korea Smart Farm R\&D Foundation (KosFarm) through the Smart Farm Innovation Technology Development Program, funded by the Ministry of Agriculture, Food and Rural Affairs (MAFRA), Ministry of Science and ICT (MSIT), and Rural Development Administration (RDA) (No. 421027-04).

\bibliography{mybibfile}

\end{document}